\documentclass[10pt,twocolumn]{article}

\usepackage{cvpr}
\usepackage{times}
\usepackage{epsfig}
\usepackage{graphicx}
\usepackage{amsmath}
\usepackage{amssymb}
\usepackage{multirow}
\usepackage{array}

\usepackage{amsthm}
\usepackage{xspace}
\usepackage{xcolor}

\usepackage{enumitem} 
\usepackage{colortbl}
\usepackage{contour}
\usepackage{nicefrac}

\usepackage{pgfplots}
\usepackage{pgfplotstable}
\usepackage{pgf,tikz}

\newcommand{\extdata}[1]{\input{#1}}

\usepackage[pagebackref=true,breaklinks=true,letterpaper=true,colorlinks,bookmarks=false]{hyperref}

\cvprfinalcopy 

\ifcvprfinal\fi
\begin{document}

\title{Revisiting Oxford and Paris: Large-Scale Image Retrieval Benchmarking}

\author{
Filip Radenovi{\'c}$^1$ \ \ \ \ Ahmet Iscen$^1$ \ \ \ \ Giorgos Tolias$^1$\ \ \ \ Yannis Avrithis$^2$\ \ \ \ Ond{\v r}ej Chum$^{1}$\\
{\fontsize{11}{13}\selectfont$^1$VRG, FEE, CTU in Prague\ \ \ \ \ \ $^2$Inria Rennes}\\
}

\maketitle
\thispagestyle{empty}

\newcommand\blfootnote[1]{%
  \begingroup
  \renewcommand\thefootnote{}\footnote{#1}%
  \addtocounter{footnote}{-1}%
  \endgroup
}
\blfootnote{The authors were supported by the MSMT LL1303 ERC-CZ grant.}

\xspaceaddexceptions{+()}

\newcommand{\real}{\mathbb{R}}
\newcommand{\integer}{\mathbb{Z}}
\def\l2{\ensuremath{\ell_2}\xspace}
\def\linf{\ensuremath{\ell_\infty}\xspace}

\def\p{\ensuremath{p}\xspace}
\def\cI{\mathcal{I}}
\def\cS{\mathcal{S}}
\def\y{\mathbf{y}}
\def\x{\mathbf{x}}

\def\sssp{\hspace{1pt}}
\def\ssp{\hspace{3pt}}
\def\msp{\hspace{5pt}}
\def\bsp{\hspace{12pt}}

\def\etal{\emph{et al.}\xspace}
\def\ie{\emph{i.e.}\xspace}
\def\eg{\emph{e.g.}\xspace}

\def\vlad{VLAD\xspace}
\def\smk{SMK$^{\star}$\xspace}
\def\asmk{ASMK$^{\star}$\xspace}
\def\sp{SP\xspace}
\def\qe{QE\xspace}
\def\hqe{HQE\xspace}
\def\dfs{DFS\xspace}
\def\off{O}
\def\hesaff{HesAff\xspace}
\def\rsift{rSIFT\xspace}
\def\delf{DELF\xspace}

\def\roxf{$\mathcal{R}$Oxford\xspace}
\def\rox{$\mathcal{R}$Oxf\xspace}
\def\ro{$\mathcal{R}$O\xspace}
\def\rpar{$\mathcal{R}$Paris\xspace}
\def\rpa{$\mathcal{R}$Par\xspace}
\def\rp{$\mathcal{R}$P\xspace}
\def\r1m{$\mathcal{R}$1M\xspace}
\def\rs{$\mathcal{R}$100k\xspace}

\newcommand{\alert}[1]{{\color{red}{#1}}}
\newcommand{\gio}[1]{{\color{purple}{#1}}}
\newcommand{\fr}[1]{{\color{blue}{#1}}}
\newcommand{\och}[1]{{\color{brown}{#1}}}
\newcommand{\ahmet}[1]{{\color{orange}{#1}}}

\newcommand{\comment}[1]{}

\renewcommand{\paragraph}[1]{\vspace{.4\baselineskip}\noindent{\bf #1}\xspace}
\newcommand{\paragraphitem}[1]{\vspace{.0\baselineskip}\noindent{\underline{\textit{#1}}}\xspace}

\newcommand{\xcaption}[2][1]{\caption{#2}\vspace{-#1\baselineskip}}

\newcommand{\eq}{eqn.\xspace}

\newcommand{\our}{$\boldsymbol{\star}$\xspace}

\definecolor{greenn}{rgb}{0.30,0.69,0.31}

\definecolor{sota}{rgb}{0.298,0.686,0.314}
\definecolor{sotaqe}{rgb}{1.0000,0.5961,0}

\definecolor{query}{rgb}{0.1020,0.1373,0.4941}
\definecolor{easy}{rgb}{0.1059,0.3686,0.1255}
\definecolor{hard}{rgb}{0.5451,0.7647,0.2902}
\definecolor{unclear}{rgb}{1.0000,0.8392,0}
\definecolor{negative}{rgb}{0.7765,0.1569,0.0157}

\definecolor{ba}{rgb}{0.8980, 0.4510, 0.4510}
\definecolor{bA}{rgb}{0.7176, 0.1098, 0.1098}

\definecolor{bb}{rgb}{0.9412, 0.3843, 0.5725}
\definecolor{bB}{rgb}{0.5333, 0.0549, 0.3098}

\definecolor{bc}{rgb}{0.7294, 0.4078, 0.7843}
\definecolor{bC}{rgb}{0.2902, 0.0784, 0.5490}

\definecolor{bd}{rgb}{0.3922, 0.7098, 0.9647}
\definecolor{bD}{rgb}{0.0510, 0.2784, 0.6314}

\definecolor{be}{rgb}{0.3020, 0.8157, 0.8824}
\definecolor{bE}{rgb}{0.0000, 0.3765, 0.3922}

\definecolor{bf}{rgb}{0.5059, 0.7804, 0.5176}
\definecolor{bF}{rgb}{0.1059, 0.3686, 0.1255}

\definecolor{bg}{rgb}{1.0000, 0.9451, 0.4627}
\definecolor{bG}{rgb}{0.9608, 0.4980, 0.0902}

\definecolor{bh}{rgb}{0.6314, 0.5333, 0.4980}
\definecolor{bH}{rgb}{0.3059, 0.2039, 0.1804}

\newcommand{\soaf}[1]{{\textbf{\color{red}{#1}}}}
\newcommand{\soas}[1]{{\textbf{\color{black}{#1}}}}
\newcommand{\soat}[1]{{\textbf{\color{blue}{#1}}}}

\newenvironment{itemizes}{%
\begin{list} {$\bullet$} {\setlength{\itemsep}{0pt}
                          \setlength{\leftmargin}{15pt}
                          \setlength{\topsep}{3pt} } }
{\end{list}\vspace{5pt}}

\newcommand{\mypar}[1]{\noindent \textbf{#1}}

\newcommand{\head}[1]{{\smallskip\noindent\bf #1}}

\newcommand{\ochout}[1]{}

\begin{abstract}
In this paper we address issues with image retrieval benchmarking on standard and popular Oxford~5k and Paris~6k datasets. In particular, annotation errors, the size of the dataset, and the level of challenge are addressed: new annotation for both datasets is created with an extra attention to the reliability of the ground truth. Three new protocols of varying difficulty are introduced. The protocols allow fair comparison between different methods, including those using a dataset pre-processing stage.
For each dataset, 15 new challenging queries are introduced. Finally, a new set of 1M hard, semi-automatically cleaned distractors is selected.

An extensive\footnote{We thank Facebook for the donation of GPU servers, which made the evaluation tractable.} comparison of the state-of-the-art methods is performed on the new benchmark.
Different types of methods are evaluated, ranging from local-feature-based to modern CNN based methods.
The best results are achieved by taking the best of the two worlds. Most importantly, image retrieval appears far from being solved.

\end{abstract}

\vspace{-10pt}
\section{Introduction}
\label{sec:introduction}

Image retrieval methods have gone through significant development in the last decade, starting with descriptors based on local-features, first organized in bag-of-words~\cite{SZ03}, and further expanded by spatial verification~\cite{PCISZ07}, hamming embedding~\cite{JDS08}, and query expansion~\cite{CPSIZ07}. Compact representations reducing the memory footprint and speeding up queries started with aggregating local descriptors~\cite{JDSP10}. Nowadays, the most efficient retrieval methods are based on fine-tuned convolutional neural networks (CNNs)~\cite{GARL17,RTC17,NAS+17}.

In order to measure the progress and compare different methods, standardized image retrieval benchmarks are used. Besides the fact that a benchmark should simulate a real-world application, there are a number of properties that determine the quality of a benchmark: the \emph{reliability of the annotation}, the \emph{size}, and the \emph{challenge level}.

Errors in the annotation may systematically corrupt the comparison of different methods. Too small datasets are prone to over-fitting and do not allow the evaluation of the efficiency of the methods. The reliability of the annotation and size of the dataset are competing factors, as it is difficult to secure accurate human annotation of large datasets. The size is commonly increased by adding a distractor set, which contains irrelevant images that are selected in an automated manner (different tags, GPS information, \etc) Finally, benchmarks where all the methods achieve almost perfect results~\cite{LBBH98} cannot be used for further improvement or qualitative comparison.

Many datasets have been introduced to measure the performance of image retrieval.
Oxford~\cite{PCISZ07}  and Paris~\cite{PCISZ08} datasets belong to the most popular ones. Numerous methods of image retrieval~\cite{CPSIZ07,PCM09,CMPM11,MPCM13,TJ14,BL15,TSJ16,KMO15,RTC17,GARL17} and visual localization~\cite{GBQV09,AGTPS15} have used these datasets for evaluation. One reason for their popularity is that, in contrast to datasets that contain small groups of 4-5 similar images like Holidays~\cite{JDS08} and UKBench~\cite{NiS06}, Oxford and Paris contain queries with up to hundreds of positive images.

Despite the popularity, there are known issues with the two datasets, which are related to all three important properties of evaluation benchmarks. First, there are errors in the annotation, including both false positives and false negatives. Further inaccuracy is introduced by queries of different sides of a landmark, sharing the annotation despite being visually distinguishable. Second, the annotated datasets are relatively small (5,062 and 6,392 images respectively). Third, current methods report near-perfect results on both the datasets. It has become difficult to draw conclusions from quantitative evaluations, especially given the annotation errors~\cite{ITAFC17}.

The lack of difficulty is not caused by the fact that non-trivial instances are not present in the dataset, but due to the annotation. The annotation was introduced about ten years ago.
At that time, the annotators had different perception of what the limits of image retrieval are. Many instances that are nowadays considered as a change of viewpoint expected to be retrieved, are \emph{de facto} excluded from the evaluation by being labelled as \emph{Junk}.

\pagebreak
The size issue of the datasets is partially addressed by the Oxford~100k \emph{distractor set}. However, this contains false negative images, as well as images that are not challenging. State-of-the-art methods maintain near-perfect results even in the presence of these distractors. As a result, additional computational effort is spent with little benefit in drawing conclusions.

\paragraph{Contributions.}
As a first contribution, we generate new annotation for Oxford and Paris datasets, update the evaluation protocol, define new, more difficult queries, and create new set of challenging distractors.
As an outcome we produce \emph{Revisited Oxford}, \emph{Revisited Paris}, and an accompanying distractor set of one million images. We refer to them as \roxf, \rpar, and \r1m respectively.

As a second contribution, we provide extensive evaluation of image retrieval methods, ranging from local-feature based to CNN-descriptor based approaches, including various methods of re-ranking.
\section{Revisiting the datasets}
\label{sec:revisit}
In this section we describe in detail why and how we revisit the annotation of Oxford and Paris datasets, present a new evaluation protocol and an accompanying challenging set of one million distractor images.
The revisited benchmark is publicly available\footnote{\url{cmp.felk.cvut.cz/revisitop}}.

\begin{figure*}[t]
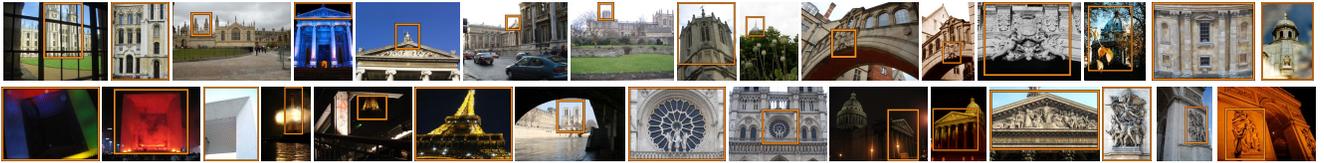

\vspace{-15pt}
\begin{center}
\foreach \q in {56,...,58}
{\includegraphics[height=1.03cm]{fig/oxford5k_queries/q\q.jpg} \hspace{-1pt}}%
\foreach \q in {59,...,61}
{\includegraphics[height=1.03cm]{fig/oxford5k_queries/q\q.jpg} \hspace{-1pt}}%
\foreach \q in {62,...,64}
{\includegraphics[height=1.03cm]{fig/oxford5k_queries/q\q.jpg} \hspace{-1pt}}%
\foreach \q in {65,...,67}
{\includegraphics[height=1.03cm]{fig/oxford5k_queries/q\q.jpg} \hspace{-1pt}}%
\foreach \q in {68,...,70}
{\includegraphics[height=1.03cm]{fig/oxford5k_queries/q\q.jpg} }
\\
\vspace{2pt}
\foreach \q in {56,...,58}
{\includegraphics[height=0.97cm]{fig/paris6k_queries/q\q.jpg} \hspace{-1pt}}%
\foreach \q in {59,...,61}
{\includegraphics[height=0.97cm]{fig/paris6k_queries/q\q.jpg} \hspace{-1pt}}%
\foreach \q in {62,...,64}
{\includegraphics[height=0.97cm]{fig/paris6k_queries/q\q.jpg} \hspace{-1pt}}%
\foreach \q in {65,...,67}
{\includegraphics[height=0.97cm]{fig/paris6k_queries/q\q.jpg} \hspace{-1pt}}%
\foreach \q in {68,...,70}
{\includegraphics[height=0.97cm]{fig/paris6k_queries/q\q.jpg} }
\end{center}
\vspace{-10pt}
\caption{The newly added queries for \roxf (top) and \rpar (bottom) datasets. Merged with the original queries, they comprise a new set of 70 queries in total.\label{fig:new_queries}}%
\end{figure*}

\subsection{The original datasets}

The original Oxford and Paris datasets consist of 5,063 and 6,392 high-resolution ($1024\times 768$) images, respectively.
Each dataset contains 55 queries comprising 5 queries per landmark, coming from a total of 11 landmarks.
Given a landmark query image, the goal is to retrieve all database images depicting the same landmark.
The original annotation (labeling) is performed manually and consists of 11 ground truth lists since 5 images of the same landmark form a \emph{query group}. Three labels are used, namely, \emph{positive}, \emph{junk}, and \emph{negative}~\footnote{We rename the originally used labels \{good, ok, junk, and absent\} for the purpose of consistency with our terminology. Good and ok were always used as positives.}.

Positive images clearly depict more than 25\% of the landmark, junk less than 25\%, while the landmark is not shown in negative ones.
The performance is measured via mean average precision (mAP)~\cite{PCISZ07} over all 55 queries, while junk images are ignored,
\ie the evaluation is performed as if they were not present in the database.

\subsection{Revisiting the annotation}
The annotation is performed by five annotators, and it is performed in the following steps.

\paragraph{Query groups.}
Query groups share the same ground-truth list and simplify the labeling problem, but also cause some inaccuracies in the original annotation.
\textit{Balliol} and \textit{Christ Church} landmarks are depicted from a different (not fully symmetric) side in the $2^\text{nd}$ and $4^\text{th}$ query, respectively. \textit{Arc de Triomphe} has three day and two night queries, while day-night matching is considered a challenging problem~\cite{VYFL15,RSJFCM16}.
We alleviate this by splitting these cases into separate groups. As a result, we form 13 and 12 query groups on Oxford and Paris, respectively.

\begin{figure*}[t!]
\vspace{-10pt}
\newcommand\hh{1cm}
\setlength{\fboxsep}{0pt} \setlength{\fboxrule}{2pt}
\begin{center}
\hspace{2mm}%
\fcolorbox{query}{black}{\includegraphics[height=\hh]{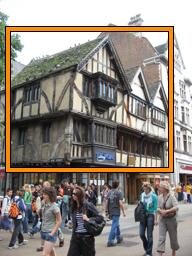}} %
\raisebox{10pt}{\hspace{-2pt}{\large$\rightarrow$}\hspace{-2pt}}
\foreach \db in {1626}%
{%
	\fcolorbox{negative}{black}{\includegraphics[height=\hh]{fig/oxford5k_mistakes/n2e_q26_db\db.jpg}} %
}%
\hspace{2mm}%
\fcolorbox{query}{black}{\includegraphics[height=\hh]{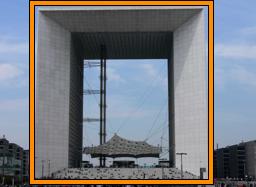}} %
\raisebox{10pt}{\hspace{-2pt}{\large$\rightarrow$}\hspace{-2pt}}
\foreach \db in {212,2705,6074}%
{%
	\fcolorbox{negative}{black}{\includegraphics[height=\hh]{fig/paris6k_mistakes/n2e_q1_db\db.jpg}} %
}%
\hspace{2mm}%
\fcolorbox{query}{black}{\includegraphics[height=\hh]{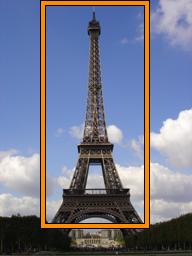}} %
\raisebox{10pt}{\hspace{-2pt}{\large$\rightarrow$}\hspace{-2pt}}
\foreach \db in {813,1787,3082,5103,5117,6220}%
{%
	\fcolorbox{negative}{black}{\includegraphics[height=\hh]{fig/paris6k_mistakes/n2e_q6_db\db.jpg}} %
}%
\hspace{2mm}%
\fcolorbox{query}{black}{\includegraphics[height=\hh]{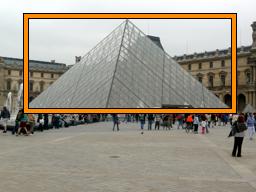}} %
\raisebox{10pt}{\hspace{-2pt}{\large$\rightarrow$}\hspace{-2pt}}
\foreach \db in {890,1689,2210,2674}%
{%
	\fcolorbox{negative}{black}{\includegraphics[height=\hh]{fig/paris6k_mistakes/n2e_q16_db\db.jpg}} %
}%
\hspace{2mm}%
\fcolorbox{query}{black}{\includegraphics[height=\hh]{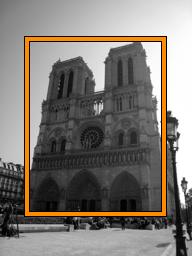}} %
\raisebox{10pt}{\hspace{-2pt}{\large$\rightarrow$}\hspace{-2pt}}
\foreach \db in {5769,6206}%
{%
	\fcolorbox{negative}{black}{\includegraphics[height=\hh]{fig/paris6k_mistakes/n2e_q31_db\db.jpg}} %
}%
\hspace{2mm}%
\fcolorbox{query}{black}{\includegraphics[height=\hh]{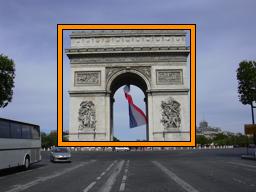}} %
\raisebox{10pt}{\hspace{-2pt}{\large$\rightarrow$}\hspace{-2pt}}
\foreach \db in {4774}%
{%
	\fcolorbox{negative}{black}{\includegraphics[height=\hh]{fig/paris6k_mistakes/n2e_q51_db\db.jpg}} %
}%
\\
\hspace{2mm}%
\fcolorbox{query}{black}{\includegraphics[height=\hh]{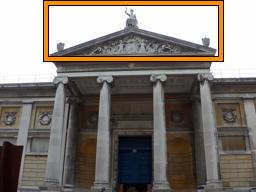}} %
\raisebox{10pt}{\hspace{-2pt}{\large$\rightarrow$}\hspace{-2pt}}
\foreach \db in {919}%
{%
	\fcolorbox{easy}{black}{\includegraphics[height=\hh]{fig/oxford5k_mistakes/o2n_q6_db\db.jpg}} %
}%
\hspace{2mm}%
\fcolorbox{query}{black}{\includegraphics[height=\hh]{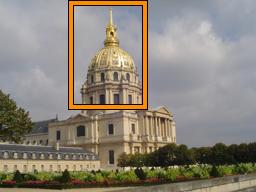}} %
\raisebox{10pt}{\hspace{-2pt}{\large$\rightarrow$}\hspace{-2pt}}
\foreach \db in {1751,3603,5142,5260,6310}%
{%
	\fcolorbox{easy}{black}{\includegraphics[height=\hh]{fig/paris6k_mistakes/o2n_q11_db\db.jpg}} %
}%
\hspace{2mm}%
\fcolorbox{query}{black}{\includegraphics[height=\hh]{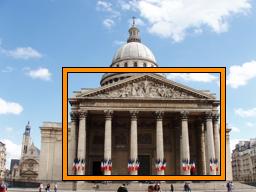}} %
\raisebox{10pt}{\hspace{-2pt}{\large$\rightarrow$}\hspace{-2pt}}
\foreach \db in {5312}%
{%
	\fcolorbox{easy}{black}{\includegraphics[height=\hh]{fig/paris6k_mistakes/o2n_q36_db\db.jpg}} %
}%
\end{center}
\vspace{-10pt}
\caption{Examples of \emph{extreme} labeling mistakes in the original labeling. We show the {\color{query}{\textbf{query (blue)}}} image and the associated database images that were originally marked as {\color{negative}{\textbf{negative (red)}}} or {\color{easy}{\textbf{positive (green)}}}. Best viewed in color.\label{fig:mistakes}\vspace{-15pt}}
\end{figure*}

\paragraph{Additional queries.}
We introduce new and more challenging queries (see Figure~\ref{fig:new_queries}) compared to the original ones. There are 15 new queries per dataset, originating from five out of the original 11 landmarks, with three queries per landmark. Along with the 55 original queries, they comprise the new set of 70 queries per dataset. The query groups, defined by visual similarity, are 26 and 25 for \roxf and \rpar, respectively.
As in the original datasets, the query object bounding boxes are simulating not only a user attempting to remove background clutter, but also cases of large \mbox{occlusion}.

\begin{table}[b]
\vspace{-5pt}
\newcolumntype{L}[1]{>{\raggedright\let\newline\\\arraybackslash\hspace{0pt}}m{#1}}
\newcolumntype{C}[1]{>{\centering\let\newline\\\arraybackslash\hspace{0pt}}m{#1}}
\newcolumntype{R}[1]{>{\raggedleft\let\newline\\\arraybackslash\hspace{0pt}}m{#1}}
\newcommand\cw{0.7cm}
\def\arraystretch{1.1}
\scriptsize
\setlength{\tabcolsep}{0mm}

\begin{center}
\begin{tabular}{|@{~}L{1cm}|C{\cw}|C{\cw}|C{\cw}|C{\cw}|}
    \hline
    \multicolumn{5}{|c|}{\roxf}\\ \hline 
    Labels & Easy & Hard & Uncl. & Neg. \\
    \hline
    Positive & 438 & 50 & 93 & 1 \\
    Junk & 50 & 222 & 72 & 9 \\
    Negative & 1 & 72 & 133 & 63768 \\
    \hline
\end{tabular}%
\hspace{2mm}%
\setlength{\tabcolsep}{0mm}
\begin{tabular}{|@{~}L{1cm}|C{\cw}|C{\cw}|C{\cw}|C{\cw}|}
    \hline
    \multicolumn{5}{|c|}{\rpar}\\ \hline 
    Labels & Easy & Hard & Uncl. & Neg. \\
    \hline
    Positive & 1222 & 643 & 136 & 6 \\
    Junk & 91 & 813 & 835 & 61 \\
    Negative & 16 & 147 & 273 & 71621 \\
    \hline
\end{tabular}
\end{center}
\vspace{-7pt}
\caption{Number of images switching their labeling from the original annotation (positive, junk, negative) to the new one (easy, hard, unclear, negative).
\label{tab:gnd_old_vs_new}
}
\end{table}

\paragraph{Labeling step 1: Selection of potential positives.}
Each annotator manually inspects the whole dataset and marks images depicting any side or version of a landmark. The goal is to collect all images  that are originally incorrectly labeled as negative. Even uncertain cases are included in this step and the process is repeated for each landmark.
Apart from inspecting the whole dataset, an interactive retrieval tool is used to actively search for further possible positive images.
All images marked in this phase are merged together with images originally annotated as positive or junk, creating a list of \emph{potential positives} for each landmark.

\paragraph{Labeling step 2: Label assignment.}
In this step, each annotator manually inspects the list of potential positives for each query group and assigns labels.
The possible labels are \textit{Easy}, \textit{Hard}, \textit{Unclear}, and \textit{Negative}. All images not in the list of potential positives are automatically marked negative. The instructions given to the annotators for each of the labels are as follows.
\vspace{-5pt}
\begin{itemize}[leftmargin=*]
\setlength\itemsep{-5pt}
  \item \textit{Easy:} The image clearly depicts the query landmark from the same side, with no large viewpoint change, no significant occlusion, no extreme illumination change, and no severe background clutter. In the case of fully symmetric sides, any side is valid.
  \item \textit{Hard:} The image depicts the query landmark, but with viewing conditions that are difficult to match with the query. The depicted (side of the) landmark is recognizable without any contextual visual information.
  \item \textit{Unclear:} (a) The image possibly depicts the landmark in question, but the content is not enough to make a certain guess about the overlap with the query region, or context is needed to clarify.
  (b) The image depicts a different side of a partially symmetric building, where the symmetry is significant and discriminative enough.
  \item \textit{Negative:} The image is not satisfying any of the previous conditions. For instance, it depicts a different side of the landmark compared to that of the query, with no discriminative symmetries. If the image has any physical overlap with the query, it is never negative, but rather unclear, easy, or hard according to the above.
\end{itemize}

\begin{figure*}
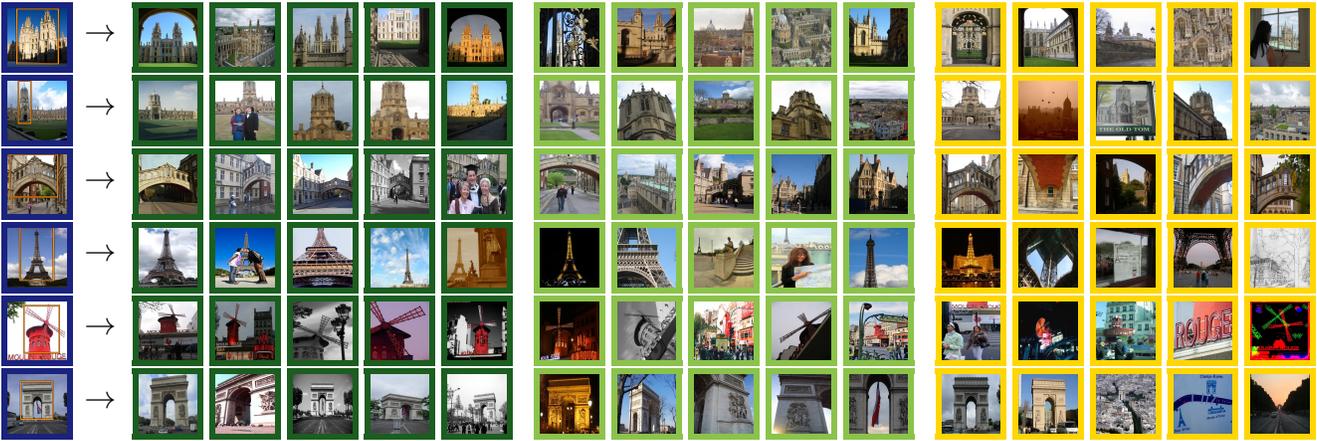

\vspace{-20pt}
\newcommand\hh{0.8cm}
\setlength{\fboxsep}{0pt} \setlength{\fboxrule}{2pt}
\begin{center}
\foreach \q in {1,21,31}
{
	\fcolorbox{query}{black}{\includegraphics[height=\hh, width=\hh]{fig/oxford5k_queries/q\q.jpg}} %
	\raisebox{10pt}{\hspace{2pt}{\large$\rightarrow$}\hspace{-2pt}}
	\hspace{2mm}%
	\foreach \e in {1,...,5}%
	{%
		\fcolorbox{easy}{black}{\includegraphics[height=\hh, width=\hh]{fig/oxford5k_labels/q\q_e\e.jpg}} %
	}%
	\hspace{2mm}%
	\foreach \h in {1,...,5}%
	{%
		\fcolorbox{hard}{black}{\includegraphics[height=\hh, width=\hh]{fig/oxford5k_labels/q\q_h\h.jpg}} %
	}%
	\hspace{2mm}%
	\foreach \u in {1,...,5}%
	{%
		\fcolorbox{unclear}{black}{\includegraphics[height=\hh, width=\hh]{fig/oxford5k_labels/q\q_u\u.jpg}} %
	}%
	\\%
}
\foreach \q in {6,25,51}
{
	\fcolorbox{query}{black}{\includegraphics[height=\hh, width=\hh]{fig/paris6k_queries/q\q.jpg}} %
	\raisebox{10pt}{\hspace{2pt}{\large$\rightarrow$}\hspace{-2pt}}
	\hspace{2mm}%
	\foreach \e in {1,...,5}%
	{%
		\fcolorbox{easy}{black}{\includegraphics[height=\hh, width=\hh]{fig/paris6k_labels/q\q_e\e.jpg}} %
	}%
	\hspace{2mm}%
	\foreach \h in {1,...,5}%
	{%
		\fcolorbox{hard}{black}{\includegraphics[height=\hh, width=\hh]{fig/paris6k_labels/q\q_h\h.jpg}} %
	}%
	\hspace{2mm}%
	\foreach \u in {1,...,5}%
	{%
		\fcolorbox{unclear}{black}{\includegraphics[height=\hh, width=\hh]{fig/paris6k_labels/q\q_u\u.jpg}} %
	}%
	\\%
}
\end{center}
\vspace{-10pt}
\caption{Sample {\color{query}{\textbf{query (blue)}}} images and images that are respectively marked as {\color{easy}{\textbf{easy (dark green)}}}, {\color{hard}{\textbf{hard (light green)}}}, and {\color{unclear}{\textbf{unclear (yellow)}}}. Best viewed in color.\label{fig:labels}\vspace{-15pt}}
\end{figure*}

\paragraph{Labeling step 3: Refinement.}
For each query group, each image in the list of potential positives has been assigned a five-tuple of labels, one per annotator. We perform majority voting in two steps to define the final label. The first step is voting for \{easy,hard\}, \{unclear\}, or \{negative\}, grouping easy and hard together. In case majority goes to \{easy,hard\}, the second step is to decide which of the two.
Draws of the first step are assigned to unclear, and of the second step to hard.
Illustrative examples are (EEHUU) $\rightarrow$ E, (EHUUN) $\rightarrow$ U, and (HHUNN) $\rightarrow$ U.
Finally, for each query group, we inspect images by descending label entropy to make sure there are no errors.
\paragraph{Revisited datasets: \roxf and \rpar.}
Images from which the queries are cropped are excluded from the evaluation dataset. This way, unfair comparisons are avoided in the case of methods performing off-line preprocessing of the database~\cite{AZ12,ITAFC17}; any preprocessing should not include any part of query images. The revisited datasets, namely, \roxf and \rpar, comprise 4,993 and 6,322 images respectively, after removing the 70 queries.

In Table~\ref{tab:gnd_old_vs_new}, we show statistics of label transitions from the old to the new annotations. Note that errors in the original annotation that affect the evaluation, \eg negative moving to easy or hard, are not uncommon. The transitions from junk to easy or hard are reflecting the greater challenges of the new annotation. Representative examples of \emph{extreme} labeling errors of the original annotation are shown in Figure~\ref{fig:mistakes}. In Figure~\ref{fig:labels}, representative examples of easy, hard, and unclear images are presented for several queries. This will help understanding the level of challenge of each evaluation protocol listed below.

\subsection{Evaluation protocol}
Only the cropped regions are to be used as queries; never the full image, since the ground-truth labeling strictly considers only the visual content inside the query region.

The standard practice of reporting mean average precision (mAP)~\cite{PCISZ07} for performance evaluation is followed. Additionally, mean precision at rank $K$ (mP@$K$) is reported. The former reflects the overall quality of the ranked list. The latter reflects the quality of the results of a search engine as they would be visually inspected by a user. More importantly, it is correlated to performance of subsequent processing steps~\cite{CPSIZ07,KTAP+11}.
During the evaluation, positive images should be retrieved, while there is also an ignore list per query. Three evaluation setups of different difficulty are defined by treating labels (easy, hard, unclear) as positive or negative, or ignoring them:
\begin{itemize}[leftmargin=*]
\setlength\itemsep{-15pt}
\item \textbf{Easy (E):} \textit{Easy} images are treated as positive, while \textit{Hard} and \textit{Unclear} are ignored (same as \textit{Junk} in~\cite{PCISZ07}).\\
\item \textbf{Medium (M):} \textit{Easy} and \textit{Hard} images are treated as positive, while \textit{Unclear} are ignored.\\
\item \textbf{Hard (H):} \textit{Hard} images are treated as positive, while \textit{Easy} and \textit{Unclear} are ignored.
\end{itemize}

If there are no positive images for a query in a particular setting, then that query is excluded from the evaluation.

The original annotation and evaluation protocol is closest to our \textbf{Easy} setup. Even though this setup is now trivial for the best performing methods, it can still be used for evaluation of \eg near duplicate detection or retrieval with ultra short codes. The other setups, \textbf{Medium} and \textbf{Hard}, are challenging and even the best performing methods achieve relatively low scores. See Section~\ref{sec:experiments} for details.

\begin{figure}[t]
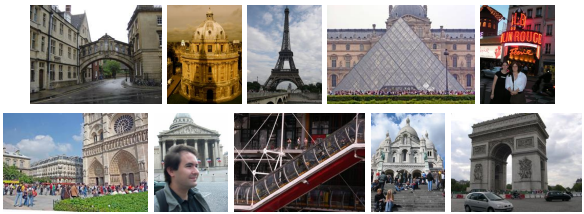

\vspace{10pt}
\begin{center}
\foreach \m in {1,...,5}%
{%
	\includegraphics[height=1.3cm]{fig/oxford100k_mistakes/m\m.jpg} %
}%
\\\vspace{1mm}%
\foreach \m in {6,...,10}%
{%
	\includegraphics[height=1.3cm]{fig/oxford100k_mistakes/m\m.jpg} %
}%
\end{center}
\vspace{-10pt}
\caption{Sample false negative images in Oxford100k.\label{fig:ox100k_mistakes}\vspace{-10pt}}%
\end{figure}
\begin{figure*}[t]
\vspace{-15pt}
\begin{center}
\includegraphics[height=1.3cm, width=1.3cm]{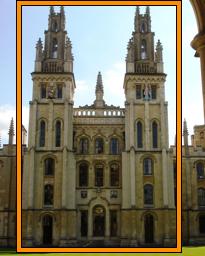} %
\raisebox{15pt}{$\rightarrow$}
\foreach \h in {1,...,10}%
{%
	\includegraphics[height=1.3cm, width=1.3cm]{fig/oxford5k_hardest/q2_h\h.jpg} %
}%
\\\vspace{1mm}%
\includegraphics[height=1.3cm, width=1.3cm]{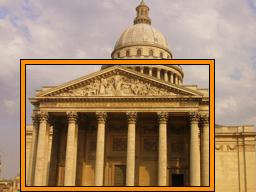} %
\raisebox{15pt}{$\rightarrow$}
\foreach \h in {1,...,10}%
{%
	\includegraphics[height=1.3cm, width=1.3cm]{fig/paris6k_hardest/q37_h\h.jpg} %
}%
\end{center}
\vspace{-10pt}
\caption{The most distracting images per query for two queries.%
\label{fig:hardest_distractors}}%
\vspace{-10pt}
\end{figure*}

\subsection{Distractor set \r1m}

Large scale experiments on Oxford and Paris dataset are commonly performed with the accompanying distractor set of 100k images, namely Oxford100k~\cite{PCISZ07}. Recent results~\cite{ITAFC17,IATFC18} show that the performance only slightly degrades by adding Oxford100k in the database compared to a small-scale setting. Moreover, it is not manually cleaned and, as a consequence, Oxford and Paris landmarks are depicted in some of the distractor images (see Figure~\ref{fig:ox100k_mistakes}), hence adding further noise to the evaluation procedure.

Larger distactor sets are used in the literature~\cite{PCISZ07,PCISZ08,JDS08,TA11} but none of them are standardized to provide a testbed for direct large scale comparison nor are they manually cleaned~\cite{JDS08}. Some of the distractor sets are also biased, since they contain images of different resolution than the Oxford and Paris datasets.

We construct a new distractor set with exactly 1,001,001 high-resolution (1024 $\times$ 768) images, which we refer to as \r1m dataset. It is cleaned by a semi-automatic process. We automatically pick hard images for a number of state-of-the-art methods, resulting in a challenging large scale setup.

\paragraph{YFCC100M and semi-automatic cleaning.}
We randomly choose 5M images with GPS information from YFCC100M dataset~\cite{TSFE+16}. Then, we exclude UK, France, and Las Vegas; the latter due to the \textit{Eiffel Tower} and \textit{Arc de Triomphe} replicas. We end up with roughly 4.1M images that are available for downloading in high resolution.
We rank images with the same search tool as used in labeling step 1. Then, we manually inspect the top 2k 
images per landmark, and remove those depicting the query landmarks (faulty GPS, toy models, and paintings/photographs of landmarks). In total, we find 110 such images.

\vspace{-5pt}
\paragraph{Un-biased mining of distracting images.}
We propose a way to keep the most challenging 1M out of the 4.1M images.
We perform all 70 queries into the 4.1M database with a number of methods. For each query and for each distractor image we count the fraction of easy or hard images that are ranked after it. We sum these fractions over all queries of \roxf and \rpar and over different methods, resulting in a measurement of how \emph{distracting} each distractor image is. We choose the set of 1M most distracting images and refer to it as the \r1m \emph{distractor set}.

Three complementary retrieval methods are chosen  to compute this measurement.
These are fine-tuned ResNet with GeM pooling~\cite{RTC17}, pre-trained (on ImageNet) AlexNet with MAC pooling~\cite{RSMC14}, and ASMK~\cite{TAJ15}. More details on these methods are given in Section~\ref{sec:baselines}. Finally, we perform a sanity check to show that this selection process is not significantly biased to distract only those 3 methods. 
This includes two additional methods, VLAD~\cite{JDSP10} and fine-tuned ResNet with R-MAC pooling by Gordo \etal~\cite{GARL17}.
As shown in Table~\ref{tab:hardest}, the performance on the hardest 1M distractors is hardly affected whether one of those additional methods participates or not in the selection process. This suggests that the mining process is not biased towards particular methods.

Table~\ref{tab:hardest} also shows that the distractor set we choose (version 1M (1,2,3) in the Table) is much harder than a random 1M subset and nearly as hard as all 4M distractor images.
Example images from the set \r1m are shown in Figure~\ref{fig:hardest_distractors}.

\begin{table}[b]
\vspace{-10pt}
\newcolumntype{L}[1]{>{\raggedright\let\newline\\\arraybackslash\hspace{0pt}}m{#1}}
\newcolumntype{C}[1]{>{\centering\let\newline\\\arraybackslash\hspace{0pt}}m{#1}}
\newcolumntype{R}[1]{>{\raggedleft\let\newline\\\arraybackslash\hspace{0pt}}m{#1}}
\def\arraystretch{1.1}
\newcommand\cw{0.6cm}
\footnotesize
\begin{center}
\resizebox{\columnwidth}{!}{%
\setlength{\tabcolsep}{0.2mm}
\begin{tabular}{|@{~}L{1.6cm}|C{\cw}|C{\cw}|C{\cw}|C{\cw}|C{\cw}|C{\cw}|C{\cw}|C{\cw}|C{\cw}|C{\cw}|}
    \hline
    \multirow{3}{*}{Distractor set} & \multicolumn{5}{c|}{\roxf} & \multicolumn{5}{c|}{\rpar} \\
    \cline{2-11}
    & \multicolumn{10}{c|}{Method} \\ \cline{2-11}
    & (1) & (2) & (3) & (4) & (5) & (1) & (2) & (3) & (4) & (5) \\
    \hline\hline
    4M     & 33.3 & 11.1 & 33.2 & 33.7 & 15.6 & 40.7 & 11.4 & 30.0 & 45.4 & 17.9 \\
    \hline
    1M (1,2,3) & 33.9 & 11.1 & 34.8 & 33.9 & 17.4 & 44.1 & 11.8 & 31.7 & 48.1 & 19.6  \\
    1M (1,2,3,4) & 33.7 & 11.1 & 34.8 & 33.8 & 17.5 & 43.8 & 11.8 & 31.8 & 47.7 & 19.7 \\
    1M (1,2,3,5) & 33.7 & 11.1 & 34.6 & 33.9 & 17.2 & 43.5 & 11.7 & 31.4 & 47.7 & 19.2 \\
    \hline
    1M (random) & 37.6 & 13.7 & 37.4 & 38.9 & 20.4 & 47.3 & 16.2 & 34.2 & 53.1 & 21.9 \\
    \hline
\end{tabular}
}
\end{center}
\vspace{-8pt}
\caption{Performance (mAP) evaluation with the Medium protocol for different distractor sets. The methods considered are (1) Fine-tuned ResNet101 with GeM pooling~\cite{RTC17}; (2) Off-the-shelf AlexNet with MAC pooling~\cite{RSMC14}; (3) \hesaff--\rsift--\asmk~\cite{TAJ15}; (4) Fine-tuned ResNet101 with R-MAC pooling~\cite{GARL17}; (5) \hesaff--\rsift--\vlad~\cite{JDSP10}. The sanity check includes evaluation for different distractor sets, \ie all, hardest subset chosen by method (1,2,3), (1,2,3,4), (1,2,4,5), and a random 1M sample. \label{tab:hardest}}
\end{table}

\section{Extensive evaluation}
\label{sec:baselines}
We evaluate a number of state-of-the-art approaches on the new benchmark and offer a rich testbed for future comparisons.
We list them in this section and they belong to two main categories, namely, classical retrieval approaches using local features and CNN-based methods producing global image descriptors.

\vspace{5pt}
\subsection{Local-feature-based methods}
\label{sec:base:local}
Methods based on local invariant features~\cite{MS05,MTSZMSKG05} and the Bag-of-Words (BoW) model~\cite{SZ03,PCISZ07,CPSIZ07,PCISZ08,CPM09,MPCM13,TJ14,CWLZ+10,ZWL+13,ZLL+10,TGS+14} were dominating the field of image retrieval until the advent of CNN-based approaches~\cite{RSMC14,BL15,TSJ16,KMO15,AGTPS15,GARL17,RTC17,MMOS+16,YYD15}.
A typical pipeline consists of invariant local feature detection~\cite{MTSZMSKG05}, local descriptor extraction~\cite{MS05}, quantization with a visual codebook~\cite{SZ03}, typically created with $k$-means, assignment of descriptors to visual words and finally descriptor aggregation in a single embedding~\cite{JPDSPS11,PLSP10} or individual feature indexing with an inverted file structure~\cite{TAJ13,PCISZ07,PCM09}. We consider state-of-the-art methods from both categories. In particular, we use up-right hessian-affine (\hesaff) features~\cite{PCM09}, RootSIFT (\rsift) descriptors~\cite{AZ12}, and create the codebooks on the landmark dataset from~\cite{RTC17}, same as the one used for the whitening of CNN-based methods.
Note that we always crop the queries according to the defined region and then perform any processing to be directly comparable to CNN-based methods.

We additionally follow the same BoW-based pipeline while replacing hessian-affine and RootSIFT with the deep local attentive features (\delf)~\cite{NAS+17}.
The default extraction approach is followed (\ie at most 1000 features per image), but we reduce the descriptor dimensionality to 128 and not to 40 to be comparable to RootSIFT.
This variant is a bridge between classical approaches and deep learning.

\vspace{-5pt}
\paragraph{\vlad.} The Vector of Locally Aggregated Descriptors~\cite{JDSP10} (\vlad) is created by first-order statistics of the local descriptors. The residual vectors between descriptors and the closest centroid are aggregated \wrt a codebook whose size is 256 in our experiments. We reduce its dimensionality down to 2048 with PCA, while square-root normalization is also used~\cite{JC12}.

\vspace{-5pt}
\paragraph{\smk.} The binarized version of the Selective Match Kernel~\cite{TAJ15} (\smk), a simple extension of the Hamming Embedding~\cite{JDS08} (HE) technique, uses an inverted file structure to separately indexes binarized residual vectors while it performs the matching with a selective monomial kernel function. The codebook size is 65,536 in our experiments, while burstiness normalization~\cite{JDS09a} is always used. Multiple assignment to three nearest words is used on the query side, while the hamming distance threshold is set to 52 out of 128 bits. The rest are the default parameters.

\vspace{-5pt}
\paragraph{\asmk.} The binarized version of the Aggregated Selective Match Kernel~\cite{TAJ15} (\asmk) is an extension of \smk that jointly encodes local descriptors that are assigned to the same visual word and handles the burstiness phenomenon. Same parametrization as \smk is used.

\vspace{-5pt}
\paragraph{\sp.} Spatial verification (\sp) is known to be crucial for particular object retrieval~\cite{PCISZ07} and is performed with the RANSAC algorithm~\cite{FB81}.
It is applied on the 100 top-ranked images, as these are formed by a first filtering step, \eg the \smk or \asmk method.
Its result is the number of inlier correspondences, which is one of the most intuitive similarity measures and allows to detect true positive images.
To assume that an image is spatially verified, we require 5 inliers with \asmk and 10 with other methods.

\vspace{-5pt}
\paragraph{\hqe.} Query expansion (QE), firstly introduced by Chum~\etal~\cite{CPSIZ07} in the visual domain, typically uses spatial verification to select true positive among the top retrieved result and issues an enhanced query including the verified images.
Hamming Query Expansion~\cite{TJ14} (\hqe) is combining QE with HE. We use same soft assignment as \smk and the default parameters.

\subsection{CNN-based global descriptor methods}
\label{sec:base:cnn}
We list different aspects of a CNN-based method for image retrieval, which we later combine to form different baselines that exist in the literature.

\vspace{5pt}
\paragraph{CNN architectures.} We include 3 highly influential CNN architectures, namely AlexNet~\cite{KSH12}, VGG-16~\cite{SZ14}, and ResNet101~\cite{HZRS16}.
They have different number of layers, complexity, and also produce descriptors of different dimensionality (256, 512, and 2048, respectively).

\vspace{5pt}
\paragraph{Pooling.}
A common practice is to consider a convolutional feature map and perform a pooling mechanism to construct a global image descriptor.
We consider max-pooling (MAC)~\cite{RSMC14,TSJ16}, sum-pooling (SPoC)~\cite{BL15}, weighted sum-pooling (CroW)~\cite{KMO15}, regional max-pooling (R-MAC)~\cite{TSJ16}, generalized mean-pooling (GeM)~\cite{RTC17}, and NetVLAD pooling~\cite{AGTPS15}.
The pooling is always applied on the last convolutional feature map.

\vspace{5pt}
\paragraph{Multi-scale.}
The input image is resized to a maximum 1024 $\times$ 1024 size. Then, three re-scaled versions with scaling factor of 1, $\nicefrac{1}{\sqrt{2}}$, and $\nicefrac{1}{2}$ are fed to the network. Finally, the resulting descriptors are combined into a single descriptor by average pooling~\cite{GARL17} for all methods, except for GeM where generalized-mean pooling is used~\cite{RTC17}.
This is shown to improve the performance of the CNN-based descriptors~\cite{GARL17,RTC17}.

\vspace{5pt}
\paragraph{Off-the-shelf \vs retrieval fine-tuning.}
Networks that are pre-trained on ImageNet~\cite{RDSK+15} (off-the-shelf) are directly applicable on image retrieval.
Moreover, we consider the following cases of fine-tuning for the task.
Radenovic~\etal~\cite{RTC16} fine-tune a network with landmarks photos using contrastive loss~\cite{HCL06}.
This is available with MAC~\cite{RTC16} and GeM pooling~\cite{RTC17}.
Similarly, Gordo~\etal~\cite{GARL17} fine-tune R-MAC pooling with landmark photos and triplet loss~\cite{WSLT+14}.
Finally, NetVLAD~\cite{AGTPS15} is fine-tuned using street-view images and GPS information.

\vspace{5pt}
\paragraph{Descriptor whitening} is known to be essential for such descriptors. We use the same landmark dataset~\cite{RTC17} to learn the whitening for all methods. We use PCA whitening~\cite{JC12,BL15} for all the off-the-shelf networks, and supervised whitening with SfM labels~\cite{MM07,RTC16} for all the fine-tuned ones. One exception is the tuning that includes the whitening in the network~\cite{GARL17}.

\vspace{5pt}
\paragraph{Query Expansion} is directly applicable on top of global CNN-based descriptors. More specifically, we use $\alpha$ query expansion~\cite{RTC17} ($\alpha$QE) and diffusion~\cite{ITAFC17} (DFS).

\section{Results}
\label{sec:experiments}
\begin{table}[t]
\newcolumntype{L}[1]{>{\raggedright\let\newline\\\arraybackslash\hspace{0pt}}m{#1}}
\newcolumntype{C}[1]{>{\centering\let\newline\\\arraybackslash\hspace{0pt}}m{#1}}
\newcolumntype{R}[1]{>{\raggedleft\let\newline\\\arraybackslash\hspace{0pt}}m{#1}}
\def\arraystretch{1.1}
\newcommand\cw{0.7cm}
\footnotesize
\begin{center}
\resizebox{\columnwidth}{!}{%
\setlength{\tabcolsep}{0.2mm}
\begin{tabular}{|@{~}L{2.7cm}|C{\cw}|C{\cw}|C{\cw}|C{\cw}|C{\cw}|C{\cw}|C{\cw}|C{\cw}|}
    \hline
    \multirow{2}{*}{Method} & \multirow{2}{*}{Oxf} & \multicolumn{3}{c|}{\roxf} & \multirow{2}{*}{Par} & \multicolumn{3}{c|}{\rpar} \\
    \cline{3-5}\cline{7-9}
    & & E & M & H & & E & M & H \\
    \hline\hline
    \hesaff--\rsift--\smk & 78.1 & 74.1 & 59.4 & 35.4 & 74.6 & 80.6 & 59.0 & 31.2 \\
    R--[\off]--R-MAC & 78.3 & 74.2 & 49.8 & 18.5 & 90.9 & 89.9 & 74.0 & 52.1 \\
    R--\cite{RTC17}--GeM & 87.8 & 84.8 & 64.7 & 38.5 & 92.7 & 92.1 & 77.2 & 56.3 \\
    R--\cite{RTC17}--GeM+\dfs & 90.0 & 86.5 & 69.8 & 40.5 & 95.3 & 93.9 & 88.9 & 78.5 \\
    \hline
\end{tabular}
}
\end{center}
\vspace{-8pt}%
\caption{Performance (mAP) on Oxford (Oxf) and Paris (Par) with the original annotation, and \roxf and \rpar with the newly proposed annotation with three different protocol setups: Easy (E), Medium (M), Hard (H).
\label{tab:map_old_vs_new}}
\end{table}

We report a performance comparison between the old and the revisited datasets.
Additionally, we provide an extensive evaluation of the state-of-the-art methods on the revisited dataset, with and without the new large-scale distractor set, setting up a testbed for future comparisons.

The evaluation includes local feature-based approaches (see Section~\ref{sec:base:local} for details and abbreviations), referred to by the combination of local feature type and representation method, \eg \hesaff--\rsift--\asmk.
CNN-based global descriptors are denoted with the following abbreviations.
Network architectures are AlexNet (A), VGG-16 (V), and ResNet101 (R).
The fine-tuning options are triplet loss with GPS guided mining~\cite{AGTPS15}, triplet loss with spatially verified positive pairs~\cite{GARL17}, contrastive loss with mining from 3D models~\cite{RTC16} and~\cite{RTC17}, and finally the off-the-shelf~[\off] networks.
Pooling approaches are as listed in Section~\ref{sec:base:cnn}.
For instance, ResNet101 with GeM pooling that is fine-tuned with contrastive loss and the training dataset by Radenovic~\etal~\cite{RTC17} is referred to as R--\cite{RTC17}--GeM.

\begin{table}
\vspace{-10pt}%
\footnotesize
\newcolumntype{L}[1]{>{\raggedright\let\newline\\\arraybackslash\hspace{0pt}}m{#1}}
\newcolumntype{C}[1]{>{\centering\let\newline\\\arraybackslash\hspace{0pt}}m{#1}}
\newcolumntype{R}[1]{>{\raggedleft\let\newline\\\arraybackslash\hspace{0pt}}m{#1}}
\def\arraystretch{0.96}
\begin{center}
\resizebox{\columnwidth}{!}{%
\setlength{\tabcolsep}{0.2mm}
\begin{tabular}{|@{~}L{3.3cm}|C{1.4cm}|C{1.6cm}|C{1.6cm}|C{1.3cm}|}
    \hline
    \multirow{3}{*}{Method} & \multirow{2}{*}{Memory} & \multicolumn{3}{c|}{Time (sec)} \\
    \cline{3-5}
    & & \multicolumn{2}{c|}{Extraction} & \multirow{2}{*}{Search} \\
    \cline{3-4}
    & (GB) & \footnotesize{GPU}  & \footnotesize{CPU} & \\
    \hline
    \hesaff--\rsift--ASMK$^{\star}$ & \multirow{2}{*}{62.0}  & \multirow{2}{*}{n/a + 0.06} & \multirow{2}{*}{1.08 + 2.35} & 0.98 \\
    \hesaff--\rsift--ASMK$^{\star}$+\sp & & & & 2.00 \\
    \delf--ASMK$^{\star}$+\sp & 10.3  & 0.41 + 0.01 & n/a + 0.54 & 0.52 \\
    A--\cite{RTC17}--GeM & 0.96  & 0.12 & 1.99 & 0.38 \\
    V--\cite{RTC17}--GeM & 1.92  & 0.23 & 31.11 & 0.56 \\
    R--\cite{RTC17}--GeM & 7.68  & 0.37 & 14.51 & 1.21 \\
    \hline
\end{tabular}
}
\end{center}
\vspace{-8pt}%
\caption{Time and memory measurements. Extraction time on a single thread GPU (Tesla P100) / CPU (Intel Xeon CPU E5-2630 v2 @ 2.60GHz) per image of size 1024x768, the memory requirements and the search time (single thread CPU) reported for the database of \roxf+\r1m images. Feature extraction + visual word assignment is reported for ASMK$^{\star}$.
SP: Geometry information is loaded from the disk and the loading time is included in search time. We did not consider geometry quantization~\cite{PCM09}.
\label{tab:time_mem}\vspace{-10pt}}
\end{table}

\begin{table*}[t]
\vspace{-15pt}%
\newcolumntype{L}[1]{>{\raggedright\let\newline\\\arraybackslash\hspace{0pt}}m{#1}}
\newcolumntype{C}[1]{>{\centering\let\newline\\\arraybackslash\hspace{0pt}}m{#1}}
\newcolumntype{R}[1]{>{\raggedleft\let\newline\\\arraybackslash\hspace{0pt}}m{#1}}
\def\arraystretch{1.13}
\newcommand\cw{0.8cm}

\def\a{28.3} \def\b{44.7} \def\c{14.1} \def\d{28.3} \def\e{43.6} \def\f{88.6} \def\g{18.7} \def\h{69.4} \def\i{8.8} \def\j{15.5} \def\k{0.9} \def\l{2.9} \def\m{17.5} \def\n{50.7} \def\o{3.3} \def\p{21.1} 
\def\A{67.8} \def\B{87.9} \def\C{53.8} \def\D{81.1} \def\E{78.9} \def\F{99.3} \def\G{57.3} \def\H{98.3} \def\I{43.1} \def\J{62.4} \def\K{31.2} \def\L{50.7} \def\M{59.4} \def\N{93.4} \def\O{28.0} \def\P{75.7}

\def\aa{51.9} \def\bb{70.3} \def\cc{30.8} \def\dd{49.7} \def\ee{68.9} \def\ff{96.7} \def\gg{44.2} \def\hh{90.1} \def\ii{21.8} \def\jj{35.2} \def\kk{5.2} \def\ll{15.9} \def\mm{44.7} \def\nn{79.9} \def\oo{20.3} \def\pp{51.4} 
\def\AA{80.2} \def\BB{93.7} \def\CC{74.9} \def\DD{88.6} \def\EE{92.5} \def\FF{98.9} \def\GG{87.5} \def\HH{98.1} \def\II{54.8} \def\JJ{70.6} \def\KK{48.7} \def\LL{65.9} \def\MM{84.0} \def\NN{98.3} \def\OO{76.0} \def\PP{96.6}

\newcommand{\col}[3]{%
	\newcommand*{\MinNumber}{#2}%
	\newcommand*{\MaxNumber}{#3}%
	\pgfmathsetmacro{\PercentColor}{100.0*(#1-\MinNumber+0.0*(\MaxNumber-\MinNumber))/(\MaxNumber-\MinNumber+0.0*(\MaxNumber-\MinNumber))}%
    \xdef\PercentColor{\PercentColor}%
    \cellcolor{sota!\PercentColor}{#1}
}

\newcommand{\colqe}[3]{%
    \newcommand*{\MinNumber}{#2}%
    \newcommand*{\MaxNumber}{#3}%
    \pgfmathsetmacro{\PercentColor}{100.0*(#1-\MinNumber+0.0*(\MaxNumber-\MinNumber))/(\MaxNumber-\MinNumber+0.0*(\MaxNumber-\MinNumber))}%
    \xdef\PercentColor{\PercentColor}%
    \cellcolor{sotaqe!\PercentColor}{#1}
}

\footnotesize
\begin{center}
\setlength\extrarowheight{-1pt}
\setlength{\tabcolsep}{0.0mm}
\resizebox{\textwidth}{!}{%
\begin{tabular}{|@{~}C{0.25cm}C{0.2cm}C{0.55cm}C{0.2cm}L{3.1cm}|C{\cw}|C{\cw}|C{\cw}|C{\cw}|C{\cw}|C{\cw}|C{\cw}|C{\cw}|C{\cw}|C{\cw}|C{\cw}|C{\cw}|C{\cw}|C{\cw}|C{\cw}|C{\cw}|}
    \hline
    \multicolumn{5}{|@{~}l|}{\multirow{3}{*}{Method}} & \multicolumn{8}{c|}{Medium} & \multicolumn{8}{c|}{Hard} \\
    \cline{6-21}
    & & & & & \multicolumn{2}{c|}{\rox} & \multicolumn{2}{c|}{\rox+\r1m} & \multicolumn{2}{c|}{\rpa} & \multicolumn{2}{c|}{\rpa+\r1m} & \multicolumn{2}{c|}{\rox} & \multicolumn{2}{c|}{\rox+\r1m} & \multicolumn{2}{c|}{\rpa} & \multicolumn{2}{c|}{\rpa+\r1m} \\
    \cline{6-21}
    & & & & & \tiny mAP & \tiny mP@10 & \tiny mAP & \tiny mP@10 & \tiny mAP & \tiny mP@10 & \tiny mAP & \tiny mP@10 & \tiny mAP & \tiny mP@10 & \tiny mAP & \tiny mP@10 & \tiny mAP & \tiny mP@10 & \tiny mAP & \tiny mP@10 \\
    \hline\hline
    \multicolumn{5}{|@{~}l|}{\hesaff--\rsift--\vlad} & \col{33.9}{\a}{\A} & \col{54.9}{\b}{\B} & \col{17.4}{\c}{\C} & \col{34.8}{\d}{\D} & \col{43.6}{\e}{\E} & \col{90.9}{\f}{\F} & \col{19.6}{\g}{\G} & \col{76.1}{\h}{\H} & \col{13.2}{\i}{\I} & \col{18.1}{\j}{\J} & \col{5.6}{\k}{\K} & \col{7.0}{\l}{\L} & \col{17.5}{\m}{\M} & \col{50.7}{\n}{\N} & \col{3.3}{\o}{\O} & \col{21.1}{\p}{\P} \\
    \multicolumn{5}{|@{~}l|}{\hesaff--\rsift--\smk} & \col{59.4}{\a}{\A} & \col{83.6}{\b}{\B} & \col{35.8}{\c}{\C} & \col{64.6}{\d}{\D} & \col{59.0}{\e}{\E} & \col{97.4}{\f}{\F} & \col{34.1}{\g}{\G} & \col{89.1}{\h}{\H} & \col{35.4}{\i}{\I} & \col{53.7}{\j}{\J} & \col{16.4}{\k}{\K} & \col{27.7}{\l}{\L} & \col{31.2}{\m}{\M} & \col{72.6}{\n}{\N} & \col{10.5}{\o}{\O} & \col{47.6}{\p}{\P} \\
    \multicolumn{5}{|@{~}l|}{\hesaff--\rsift--\asmk} & \col{60.4}{\a}{\A} & \col{85.6}{\b}{\B} & \col{45.0}{\c}{\C} & \col{76.0}{\d}{\D} & \col{61.2}{\e}{\E} & \col{97.9}{\f}{\F} & \col{42.0}{\g}{\G} & \col{95.3}{\h}{\H} & \col{36.4}{\i}{\I} & \col{56.7}{\j}{\J} & \col{25.7}{\k}{\K} & \col{42.1}{\l}{\L} & \col{34.5}{\m}{\M} & \col{80.6}{\n}{\N} & \col{16.5}{\o}{\O} & \col{63.4}{\p}{\P} \\
    \multicolumn{5}{|@{~}l|}{\hesaff--\rsift--\smk+\sp} & \col{59.8}{\a}{\A} & \col{84.3}{\b}{\B} & \col{38.1}{\c}{\C} & \col{67.1}{\d}{\D} & \col{59.2}{\e}{\E} & \col{97.4}{\f}{\F} & \col{34.5}{\g}{\G} & \col{89.3}{\h}{\H} & \col{35.8}{\i}{\I} & \col{54.0}{\j}{\J} & \col{17.7}{\k}{\K} & \col{30.3}{\l}{\L} & \col{31.3}{\m}{\M} & \col{73.6}{\n}{\N} & \col{11.0}{\o}{\O} & \col{49.1}{\p}{\P} \\
    \multicolumn{5}{|@{~}l|}{\hesaff--\rsift--\asmk+\sp} & \col{60.6}{\a}{\A} & \col{86.1}{\b}{\B} & \col{46.8}{\c}{\C} & \col{79.6}{\d}{\D} & \col{61.4}{\e}{\E} & \col{97.9}{\f}{\F} & \col{42.3}{\g}{\G} & \col{95.3}{\h}{\H} & \col{36.7}{\i}{\I} & \col{57.0}{\j}{\J} & \col{26.9}{\k}{\K} & \col{45.3}{\l}{\L} & \col{35.0}{\m}{\M} & \col{81.7}{\n}{\N} & \col{16.8}{\o}{\O} & \col{65.3}{\p}{\P} \\
    \multicolumn{5}{|@{~}l|}{\delf--\asmk+\sp} & \col{67.8}{\a}{\A} & \col{87.9}{\b}{\B} & \col{53.8}{\c}{\C} & \col{81.1}{\d}{\D} & \col{76.9}{\e}{\E} & \col{99.3}{\f}{\F} & \col{57.3}{\g}{\G} & \col{98.3}{\h}{\H} & \col{43.1}{\i}{\I} & \col{62.4}{\j}{\J} & \col{31.2}{\k}{\K} & \col{50.7}{\l}{\L} & \col{55.4}{\m}{\M} & \col{93.4}{\n}{\N} & \col{26.4}{\o}{\O} & \col{75.7}{\p}{\P} \\
    \hline
    A&--&[\off]&--&MAC & \col{28.3}{\a}{\A} & \col{44.7}{\b}{\B} & \col{14.1}{\c}{\C} & \col{28.3}{\d}{\D} & \col{47.3}{\e}{\E} & \col{88.6}{\f}{\F} & \col{18.7}{\g}{\G} & \col{69.4}{\h}{\H} & \col{8.8}{\i}{\I} & \col{15.5}{\j}{\J} & \col{3.5}{\k}{\K} & \col{5.1}{\l}{\L} & \col{23.1}{\m}{\M} & \col{61.6}{\n}{\N} & \col{4.1}{\o}{\O} & \col{29.0}{\p}{\P} \\
    A&--&[\off]&--&GeM & \col{33.8}{\a}{\A} & \col{51.2}{\b}{\B} & \col{16.3}{\c}{\C} & \col{32.4}{\d}{\D} & \col{52.7}{\e}{\E} & \col{90.1}{\f}{\F} & \col{23.8}{\g}{\G} & \col{78.1}{\h}{\H} & \col{10.4}{\i}{\I} & \col{16.7}{\j}{\J} & \col{3.9}{\k}{\K} & \col{6.3}{\l}{\L} & \col{26.0}{\m}{\M} & \col{68.0}{\n}{\N} & \col{5.5}{\o}{\O} & \col{31.6}{\p}{\P} \\
    A&--&~\cite{RTC16}&--&MAC & \col{41.3}{\a}{\A} & \col{62.1}{\b}{\B} & \col{23.9}{\c}{\C} & \col{43.0}{\d}{\D} & \col{56.4}{\e}{\E} & \col{92.9}{\f}{\F} & \col{29.6}{\g}{\G} & \col{85.4}{\h}{\H} & \col{17.8}{\i}{\I} & \col{28.2}{\j}{\J} & \col{8.4}{\k}{\K} & \col{11.9}{\l}{\L} & \col{28.7}{\m}{\M} & \col{69.3}{\n}{\N} & \col{8.5}{\o}{\O} & \col{40.9}{\p}{\P} \\
    A&--&~\cite{RTC17}&--&GeM & \col{43.3}{\a}{\A} & \col{62.1}{\b}{\B} & \col{24.2}{\c}{\C} & \col{42.8}{\d}{\D} & \col{58.0}{\e}{\E} & \col{91.6}{\f}{\F} & \col{29.9}{\g}{\G} & \col{84.6}{\h}{\H} & \col{17.1}{\i}{\I} & \col{26.2}{\j}{\J} & \col{9.4}{\k}{\K} & \col{11.9}{\l}{\L} & \col{29.7}{\m}{\M} & \col{67.6}{\n}{\N} & \col{8.4}{\o}{\O} & \col{39.6}{\p}{\P} \\
    \hline
    V&--&[\off]&--&MAC & \col{37.8}{\a}{\A} & \col{57.8}{\b}{\B} & \col{21.8}{\c}{\C} & \col{39.7}{\d}{\D} & \col{59.2}{\e}{\E} & \col{93.3}{\f}{\F} & \col{33.6}{\g}{\G} & \col{87.1}{\h}{\H} & \col{14.6}{\i}{\I} & \col{27.0}{\j}{\J} & \col{7.4}{\k}{\K} & \col{11.9}{\l}{\L} & \col{35.9}{\m}{\M} & \col{78.4}{\n}{\N} & \col{13.2}{\o}{\O} & \col{54.7}{\p}{\P} \\
    V&--&[\off]&--&SPoC & \col{38.0}{\a}{\A} & \col{54.6}{\b}{\B} & \col{17.1}{\c}{\C} & \col{33.3}{\d}{\D} & \col{59.8}{\e}{\E} & \col{93.0}{\f}{\F} & \col{30.3}{\g}{\G} & \col{83.0}{\h}{\H} & \col{11.4}{\i}{\I} & \col{20.9}{\j}{\J} & \col{0.9}{\k}{\K} & \col{2.9}{\l}{\L} & \col{32.4}{\m}{\M} & \col{69.7}{\n}{\N} & \col{7.6}{\o}{\O} & \col{30.6}{\p}{\P} \\
    V&--&[\off]&--&CroW & \col{41.4}{\a}{\A} & \col{58.8}{\b}{\B} & \col{22.5}{\c}{\C} & \col{40.5}{\d}{\D} & \col{62.9}{\e}{\E} & \col{94.4}{\f}{\F} & \col{34.1}{\g}{\G} & \col{87.1}{\h}{\H} & \col{13.9}{\i}{\I} & \col{25.7}{\j}{\J} & \col{3.0}{\k}{\K} & \col{6.6}{\l}{\L} & \col{36.9}{\m}{\M} & \col{77.9}{\n}{\N} & \col{10.3}{\o}{\O} & \col{45.1}{\p}{\P} \\
    V&--&[\off]&--&GeM & \col{40.5}{\a}{\A} & \col{60.3}{\b}{\B} & \col{25.4}{\c}{\C} & \col{45.6}{\d}{\D} & \col{63.2}{\e}{\E} & \col{94.6}{\f}{\F} & \col{37.5}{\g}{\G} & \col{88.6}{\h}{\H} & \col{15.7}{\i}{\I} & \col{28.6}{\j}{\J} & \col{7.6}{\k}{\K} & \col{12.1}{\l}{\L} & \col{38.8}{\m}{\M} & \col{79.0}{\n}{\N} & \col{14.2}{\o}{\O} & \col{55.9}{\p}{\P} \\
    V&--&[\off]&--&R-MAC & \col{42.5}{\a}{\A} & \col{62.8}{\b}{\B} & \col{21.7}{\c}{\C} & \col{40.3}{\d}{\D} & \col{66.2}{\e}{\E} & \col{95.4}{\f}{\F} & \col{39.9}{\g}{\G} & \col{88.9}{\h}{\H} & \col{12.0}{\i}{\I} & \col{26.1}{\j}{\J} & \col{1.7}{\k}{\K} & \col{5.8}{\l}{\L} & \col{40.9}{\m}{\M} & \col{77.1}{\n}{\N} & \col{14.8}{\o}{\O} & \col{54.0}{\p}{\P} \\
    V&--&~\cite{AGTPS15}&--&NetVLAD & \col{37.1}{\a}{\A} & \col{56.5}{\b}{\B} & \col{20.7}{\c}{\C} & \col{37.1}{\d}{\D} & \col{59.8}{\e}{\E} & \col{94.0}{\f}{\F} & \col{31.8}{\g}{\G} & \col{85.7}{\h}{\H} & \col{13.8}{\i}{\I} & \col{23.3}{\j}{\J} & \col{6.0}{\k}{\K} & \col{8.4}{\l}{\L} & \col{35.0}{\m}{\M} & \col{73.7}{\n}{\N} & \col{11.5}{\o}{\O} & \col{46.6}{\p}{\P} \\
    V&--&~\cite{RTC16}&--&MAC & \col{58.4}{\a}{\A} & \col{81.1}{\b}{\B} & \col{39.7}{\c}{\C} & \col{68.6}{\d}{\D} & \col{66.8}{\e}{\E} & \col{97.7}{\f}{\F} & \col{42.4}{\g}{\G} & \col{92.6}{\h}{\H} & \col{30.5}{\i}{\I} & \col{48.0}{\j}{\J} & \col{17.9}{\k}{\K} & \col{27.9}{\l}{\L} & \col{42.0}{\m}{\M} & \col{82.9}{\n}{\N} & \col{17.7}{\o}{\O} & \col{63.7}{\p}{\P} \\
    V&--&~\cite{RTC17}&--&GeM & \col{61.9}{\a}{\A} & \col{82.7}{\b}{\B} & \col{42.6}{\c}{\C} & \col{68.1}{\d}{\D} & \col{69.3}{\e}{\E} & \col{97.9}{\f}{\F} & \col{45.4}{\g}{\G} & \col{94.1}{\h}{\H} & \col{33.7}{\i}{\I} & \col{51.0}{\j}{\J} & \col{19.0}{\k}{\K} & \col{29.4}{\l}{\L} & \col{44.3}{\m}{\M} & \col{83.7}{\n}{\N} & \col{19.1}{\o}{\O} & \col{64.9}{\p}{\P} \\
    \hline
    R&--&[\off]&--&MAC & \col{41.7}{\a}{\A} & \col{65.0}{\b}{\B} & \col{24.2}{\c}{\C} & \col{43.7}{\d}{\D} & \col{66.2}{\e}{\E} & \col{96.4}{\f}{\F} & \col{40.8}{\g}{\G} & \col{93.0}{\h}{\H} & \col{18.0}{\i}{\I} & \col{32.9}{\j}{\J} & \col{5.7}{\k}{\K} & \col{14.4}{\l}{\L} & \col{44.1}{\m}{\M} & \col{86.3}{\n}{\N} & \col{18.2}{\o}{\O} & \col{67.7}{\p}{\P} \\
    R&--&[\off]&--&SPoC & \col{39.8}{\a}{\A} & \col{61.0}{\b}{\B} & \col{21.5}{\c}{\C} & \col{40.4}{\d}{\D} & \col{69.2}{\e}{\E} & \col{96.7}{\f}{\F} & \col{41.6}{\g}{\G} & \col{92.0}{\h}{\H} & \col{12.4}{\i}{\I} & \col{23.8}{\j}{\J} & \col{2.8}{\k}{\K} & \col{5.6}{\l}{\L} & \col{44.7}{\m}{\M} & \col{78.0}{\n}{\N} & \col{15.3}{\o}{\O} & \col{54.4}{\p}{\P} \\
    R&--&[\off]&--&CroW & \col{42.4}{\a}{\A} & \col{61.9}{\b}{\B} & \col{21.2}{\c}{\C} & \col{39.4}{\d}{\D} & \col{70.4}{\e}{\E} & \col{97.1}{\f}{\F} & \col{42.7}{\g}{\G} & \col{92.9}{\h}{\H} & \col{13.3}{\i}{\I} & \col{27.7}{\j}{\J} & \col{3.3}{\k}{\K} & \col{9.3}{\l}{\L} & \col{47.2}{\m}{\M} & \col{83.6}{\n}{\N} & \col{16.3}{\o}{\O} & \col{61.6}{\p}{\P} \\
    R&--&[\off]&--&GeM & \col{45.0}{\a}{\A} & \col{66.2}{\b}{\B} & \col{25.6}{\c}{\C} & \col{45.1}{\d}{\D} & \col{70.7}{\e}{\E} & \col{97.0}{\f}{\F} & \col{46.2}{\g}{\G} & \col{94.0}{\h}{\H} & \col{17.7}{\i}{\I} & \col{32.6}{\j}{\J} & \col{4.7}{\k}{\K} & \col{13.4}{\l}{\L} & \col{48.7}{\m}{\M} & \col{88.0}{\n}{\N} & \col{20.3}{\o}{\O} & \col{70.4}{\p}{\P} \\
    R&--&[\off]&--&R-MAC & \col{49.8}{\a}{\A} & \col{68.9}{\b}{\B} & \col{29.2}{\c}{\C} & \col{48.9}{\d}{\D} & \col{74.0}{\e}{\E} & \col{97.7}{\f}{\F} & \col{49.3}{\g}{\G} & \col{93.7}{\h}{\H} & \col{18.5}{\i}{\I} & \col{32.2}{\j}{\J} & \col{4.5}{\k}{\K} & \col{13.0}{\l}{\L} & \col{52.1}{\m}{\M} & \col{87.1}{\n}{\N} & \col{21.3}{\o}{\O} & \col{67.4}{\p}{\P} \\
    R&--&~\cite{RTC17}&--&GeM & \col{64.7}{\a}{\A} & \col{84.7}{\b}{\B} & \col{45.2}{\c}{\C} & \col{71.7}{\d}{\D} & \col{77.2}{\e}{\E} & \col{98.1}{\f}{\F} & \col{52.3}{\g}{\G} & \col{95.3}{\h}{\H} & \col{38.5}{\i}{\I} & \col{53.0}{\j}{\J} & \col{19.9}{\k}{\K} & \col{34.9}{\l}{\L} & \col{56.3}{\m}{\M} & \col{89.1}{\n}{\N} & \col{24.7}{\o}{\O} & \col{73.3}{\p}{\P} \\
    R&--&~\cite{GARL17}&--&R-MAC & \col{60.9}{\a}{\A} & \col{78.1}{\b}{\B} & \col{39.3}{\c}{\C} & \col{62.1}{\d}{\D} & \col{78.9}{\e}{\E} & \col{96.9}{\f}{\F} & \col{54.8}{\g}{\G} & \col{93.9}{\h}{\H} & \col{32.4}{\i}{\I} & \col{50.0}{\j}{\J} & \col{12.5}{\k}{\K} & \col{24.9}{\l}{\L} & \col{59.4}{\m}{\M} & \col{86.1}{\n}{\N} & \col{28.0}{\o}{\O} & \col{70.0}{\p}{\P} \\
    \hline\hline
    \multicolumn{21}{|c|}{Query expansion (QE) and diffusion (DFS)} \\
    \hline\hline
    \multicolumn{5}{|@{~}l|}{\hesaff--\rsift--\hqe} & \colqe{66.3}{\aa}{\AA} & \colqe{85.6}{\bb}{\BB} & \colqe{42.7}{\cc}{\CC} & \colqe{67.4}{\dd}{\DD} & \colqe{68.9}{\ee}{\EE} & \colqe{97.3}{\ff}{\FF} & \colqe{44.2}{\gg}{\GG} & \colqe{90.1}{\hh}{\HH} & \colqe{41.3}{\ii}{\II} & \colqe{60.0}{\jj}{\JJ} & \colqe{23.2}{\kk}{\KK} & \colqe{37.6}{\ll}{\LL} & \colqe{44.7}{\mm}{\MM} & \colqe{79.9}{\nn}{\NN} & \colqe{20.3}{\oo}{\OO} & \colqe{51.4}{\pp}{\PP} \\
    \multicolumn{5}{|@{~}l|}{\hesaff--\rsift--\hqe+\sp} & \colqe{71.3}{\aa}{\AA} & \colqe{88.1}{\bb}{\BB} & \colqe{52.0}{\cc}{\CC} & \colqe{76.7}{\dd}{\DD} & \colqe{70.2}{\ee}{\EE} & \colqe{98.6}{\ff}{\FF} & \colqe{46.8}{\gg}{\GG} & \colqe{93.0}{\hh}{\HH} & \colqe{49.7}{\ii}{\II} & \colqe{69.6}{\jj}{\JJ} & \colqe{29.8}{\kk}{\KK} & \colqe{50.1}{\ll}{\LL} & \colqe{45.1}{\mm}{\MM} & \colqe{83.9}{\nn}{\NN} & \colqe{21.8}{\oo}{\OO} & \colqe{61.9}{\pp}{\PP} \\
    \multicolumn{5}{|@{~}l|}{\delf--\hqe+\sp} & \colqe{73.4}{\aa}{\AA} & \colqe{88.2}{\bb}{\BB} & \colqe{60.6}{\cc}{\CC} & \colqe{79.7}{\dd}{\DD} & \colqe{84.0}{\ee}{\EE} & \colqe{98.3}{\ff}{\FF} & \colqe{65.2}{\gg}{\GG} & \colqe{96.1}{\hh}{\HH} & \colqe{50.3}{\ii}{\II} & \colqe{67.2}{\jj}{\JJ} & \colqe{37.9}{\kk}{\KK} & \colqe{56.1}{\ll}{\LL} & \colqe{69.3}{\mm}{\MM} & \colqe{93.7}{\nn}{\NN} & \colqe{35.8}{\oo}{\OO} & \colqe{69.1}{\pp}{\PP} \\
    \hline
    R&--&[\off]&--&R-MAC+$\alpha$\qe & \colqe{51.9}{\aa}{\AA} & \colqe{70.3}{\bb}{\BB} & \colqe{30.8}{\cc}{\CC} & \colqe{49.7}{\dd}{\DD} & \colqe{77.3}{\ee}{\EE} & \colqe{97.9}{\ff}{\FF} & \colqe{55.3}{\gg}{\GG} & \colqe{94.7}{\hh}{\HH} & \colqe{21.8}{\ii}{\II} & \colqe{35.2}{\jj}{\JJ} & \colqe{5.2}{\kk}{\KK} & \colqe{15.9}{\ll}{\LL} & \colqe{57.0}{\mm}{\MM} & \colqe{87.6}{\nn}{\NN} & \colqe{28.0}{\oo}{\OO} & \colqe{76.1}{\pp}{\PP} \\
    V&--&~\cite{RTC17}&--&GeM+$\alpha$\qe & \colqe{66.6}{\aa}{\AA} & \colqe{85.7}{\bb}{\BB} & \colqe{47.0}{\cc}{\CC} & \colqe{72.0}{\dd}{\DD} & \colqe{74.0}{\ee}{\EE} & \colqe{98.4}{\ff}{\FF} & \colqe{52.9}{\gg}{\GG} & \colqe{95.9}{\hh}{\HH} & \colqe{38.9}{\ii}{\II} & \colqe{57.3}{\jj}{\JJ} & \colqe{21.1}{\kk}{\KK} & \colqe{34.6}{\ll}{\LL} & \colqe{51.0}{\mm}{\MM} & \colqe{88.4}{\nn}{\NN} & \colqe{25.6}{\oo}{\OO} & \colqe{75.0}{\pp}{\PP} \\
    R&--&~\cite{RTC17}&--&GeM+$\alpha$\qe & \colqe{67.2}{\aa}{\AA} & \colqe{86.0}{\bb}{\BB} & \colqe{49.0}{\cc}{\CC} & \colqe{74.7}{\dd}{\DD} & \colqe{80.7}{\ee}{\EE} & \colqe{98.9}{\ff}{\FF} & \colqe{58.0}{\gg}{\GG} & \colqe{95.9}{\hh}{\HH} & \colqe{40.8}{\ii}{\II} & \colqe{54.9}{\jj}{\JJ} & \colqe{24.2}{\kk}{\KK} & \colqe{40.3}{\ll}{\LL} & \colqe{61.8}{\mm}{\MM} & \colqe{90.6}{\nn}{\NN} & \colqe{31.0}{\oo}{\OO} & \colqe{80.4}{\pp}{\PP} \\
    R&--&~\cite{GARL17}&--&R-MAC+$\alpha$\qe & \colqe{64.8}{\aa}{\AA} & \colqe{78.5}{\bb}{\BB} & \colqe{45.7}{\cc}{\CC} & \colqe{66.5}{\dd}{\DD} & \colqe{82.7}{\ee}{\EE} & \colqe{97.3}{\ff}{\FF} & \colqe{61.0}{\gg}{\GG} & \colqe{94.3}{\hh}{\HH} & \colqe{36.8}{\ii}{\II} & \colqe{53.3}{\jj}{\JJ} & \colqe{19.5}{\kk}{\KK} & \colqe{36.6}{\ll}{\LL} & \colqe{65.7}{\mm}{\MM} & \colqe{90.1}{\nn}{\NN} & \colqe{35.0}{\oo}{\OO} & \colqe{76.9}{\pp}{\PP} \\
    V&--&~\cite{RTC17}&--&GeM+\dfs & \colqe{69.6}{\aa}{\AA} & \colqe{84.7}{\bb}{\BB} & \colqe{60.4}{\cc}{\CC} & \colqe{79.4}{\dd}{\DD} & \colqe{85.6}{\ee}{\EE} & \colqe{97.1}{\ff}{\FF} & \colqe{80.7}{\gg}{\GG} & \colqe{97.1}{\hh}{\HH} & \colqe{41.1}{\ii}{\II} & \colqe{51.1}{\jj}{\JJ} & \colqe{33.1}{\kk}{\KK} & \colqe{49.6}{\ll}{\LL} & \colqe{73.9}{\mm}{\MM} & \colqe{93.7}{\nn}{\NN} & \colqe{65.3}{\oo}{\OO} & \colqe{93.1}{\pp}{\PP} \\
    R&--&~\cite{RTC17}&--&GeM+\dfs & \colqe{69.8}{\aa}{\AA} & \colqe{84.0}{\bb}{\BB} & \colqe{61.5}{\cc}{\CC} & \colqe{77.1}{\dd}{\DD} & \colqe{88.9}{\ee}{\EE} & \colqe{96.9}{\ff}{\FF} & \colqe{84.9}{\gg}{\GG} & \colqe{95.9}{\hh}{\HH} & \colqe{40.5}{\ii}{\II} & \colqe{54.4}{\jj}{\JJ} & \colqe{33.1}{\kk}{\KK} & \colqe{48.2}{\ll}{\LL} & \colqe{78.5}{\mm}{\MM} & \colqe{94.6}{\nn}{\NN} & \colqe{71.6}{\oo}{\OO} & \colqe{93.7}{\pp}{\PP} \\
    R&--&~\cite{GARL17}&--&R-MAC+\dfs & \colqe{69.0}{\aa}{\AA} & \colqe{82.3}{\bb}{\BB} & \colqe{56.6}{\cc}{\CC} & \colqe{68.6}{\dd}{\DD} & \colqe{89.5}{\ee}{\EE} & \colqe{96.7}{\ff}{\FF} & \colqe{83.2}{\gg}{\GG} & \colqe{93.3}{\hh}{\HH} & \colqe{44.7}{\ii}{\II} & \colqe{60.5}{\jj}{\JJ} & \colqe{28.4}{\kk}{\KK} & \colqe{43.6}{\ll}{\LL} & \colqe{80.0}{\mm}{\MM} & \colqe{94.1}{\nn}{\NN} & \colqe{70.4}{\oo}{\OO} & \colqe{89.1}{\pp}{\PP} \\
    \hline
    \multicolumn{5}{|@{~}l|}{\hesaff--\rsift--\asmk+\sp $\rightarrow$ R--~\cite{RTC17}--GeM+\dfs} & \colqe{79.1}{\aa}{\AA} & \colqe{92.6}{\bb}{\BB} & \colqe{74.3}{\cc}{\CC} & \colqe{87.9}{\dd}{\DD} & \colqe{91.0}{\ee}{\EE} & \colqe{98.3}{\ff}{\FF} & \colqe{85.9}{\gg}{\GG} & \colqe{97.1}{\hh}{\HH} & \colqe{52.7}{\ii}{\II} & \colqe{66.1}{\jj}{\JJ} & \colqe{48.7}{\kk}{\KK} & \colqe{65.9}{\ll}{\LL} & \colqe{81.0}{\mm}{\MM} & \colqe{97.9}{\nn}{\NN} & \colqe{73.2}{\oo}{\OO} & \colqe{96.6}{\pp}{\PP} \\
    \multicolumn{5}{|@{~}l@{~~}|}{\hesaff--\rsift--\asmk+\sp $\rightarrow$ R--~\cite{GARL17}--R-MAC+\dfs} & \colqe{80.2}{\aa}{\AA} & \colqe{93.7}{\bb}{\BB} & \colqe{74.9}{\cc}{\CC} & \colqe{87.9}{\dd}{\DD} & \colqe{92.5}{\ee}{\EE} & \colqe{98.7}{\ff}{\FF} & \colqe{87.5}{\gg}{\GG} & \colqe{97.1}{\hh}{\HH} & \colqe{54.8}{\ii}{\II} & \colqe{70.6}{\jj}{\JJ} & \colqe{47.5}{\kk}{\KK} & \colqe{62.4}{\ll}{\LL} & \colqe{84.0}{\mm}{\MM} & \colqe{98.3}{\nn}{\NN} & \colqe{76.0}{\oo}{\OO} & \colqe{96.3}{\pp}{\PP} \\
    \multicolumn{5}{|@{~}l|}{\delf--\asmk+\sp $\rightarrow$ R--~\cite{GARL17}--R-MAC+\dfs} & \colqe{75.0}{\aa}{\AA} & \colqe{87.9}{\bb}{\BB} & \colqe{68.7}{\cc}{\CC} & \colqe{83.6}{\dd}{\DD} & \colqe{90.5}{\ee}{\EE} & \colqe{98.0}{\ff}{\FF} & \colqe{86.6}{\gg}{\GG} & \colqe{98.1}{\hh}{\HH} & \colqe{48.3}{\ii}{\II} & \colqe{64.0}{\jj}{\JJ} & \colqe{39.4}{\kk}{\KK} & \colqe{55.7}{\ll}{\LL} & \colqe{81.2}{\mm}{\MM} & \colqe{95.6}{\nn}{\NN} & \colqe{74.2}{\oo}{\OO} & \colqe{94.6}{\pp}{\PP} \\
    \hline
\end{tabular}
}
\end{center}
\vspace{-8pt}%
\caption{Performance  evaluation (mAP, mP@10) on \roxf~(\rox) and \rpar~(\rpa) without and with \r1m distractors. We report results with the revisited annotation, using Medium and Hard evaluation protocols. We use a color-map that is normalized according to the minimum ({\color{white}{\protect\contour{black}{white}}}) and maximum ({\color{sota}{\textbf{green}}} / {\color{sotaqe}{\textbf{orange}}}) value per column.
\label{tab:sota_all} \vspace{-10pt}}
\end{table*}

\paragraph{Revisited \vs original.}
We compare the performance when evaluated on the original datasets, and the revisited annotation with the new protocols.
The results for four representative methods are presented in Table~\ref{tab:map_old_vs_new}.
The old setup appears to be close to the new \textbf{Easy} setup, while \textbf{Medium} and \textbf{Hard} appear to be more challenging.
We observe that the performance of the \textbf{Easy} setup is nearly saturated and, therefore, do not use it but only evaluate \textbf{Medium} and \textbf{Hard} setups in the subsequent experiments.

\paragraph{State of the art evaluation.}
We perform an extensive evaluation of the state-of-the-art methods for image retrieval.
We present time/memory measurements in Table~\ref{tab:time_mem} and performance results in Table~\ref{tab:sota_all}.
We additionally show the average precision (AP) per query for a set of representative methods in Figures~\ref{fig:ap_hist_oxf}~and~\ref{fig:ap_hist_par}, for \roxf and \rpar, respectively.
The representative set covers the progress of methods over time in the task of image retrieval.
In the evaluation, we observe that there is no single method achieving the highest score on every protocol per dataset.
Local-feature-based methods perform very well on \roxf, especially at large scale, achieving state-of-the-art performance, while CNN-based methods seem to dominate on \rpar.
We observe that BoW-based classical approaches are still not obsolete, but their improvement typically comes at significant additional cost.
Recent CNN-based local features, \ie \delf, reduce the number of features and improve the performance at the same time.

\pagebreak
CNN fine-tuning consistently brings improvements over the off-the-shelf networks. The new protocols make it clear that improvements are needed at larger scale and the hard setup.
Many images are not retrieved, while the top 10 results mostly contain false positives.
Interestingly, we observe that query expansion approaches (\eg diffusion) degrade the performance of queries with few relevant images (see Figures~\ref{fig:ap_hist_oxf}~and~\ref{fig:ap_hist_par}).
This phenomenon is more pronounced in the revisited datasets, where the the query images are removed from the preprocessing.
We did not include separate regional representation and indexing~\cite{RSMC14}, which is previously shown to be beneficial. Preliminary experiments with ResNet and GeM pooling show that it does not deliver improvements that are significant enough to justify the additional memory and complexity cost.

\begin{figure*}[t]
\vspace{-20pt}%
\input{fig_ap_hist_oxf}%
\caption{Performance (AP) per query on \roxf~+~\r1m with Medium setup. AP is shown with a bar for 8 methods. The methods, from left to right, are {\color{ba}{\protect\contour{bA}{\hesaff--\rsift--\asmk+\sp}}}, {\color{bb}{\protect\contour{bB}{\delf--\asmk+\sp}}}, {\color{bc}{\protect\contour{bC}{\delf--\hqe+\sp}}}, {\color{bd}{\protect\contour{bD}{V--[O]--R-MAC}}}, {\color{be}{\protect\contour{bE}{R--[O]--GeM}}}, {\color{bf}{\protect\contour{bF}{R--\cite{RTC17}--GeM}}}, {\color{bg}{\protect\contour{bG}{R--\cite{RTC17}--GeM+\dfs}}}, {\color{bh}{\protect\contour{bH}{\hesaff--\rsift--\asmk+\sp $\rightarrow$ R--\cite{RTC17}--GeM+\dfs}}}. The total number of easy and hard images is printed on each histogram. Best viewed in color. \label{fig:ap_hist_oxf}\vspace{-2mm}}%
\end{figure*}
\begin{figure*}[t!]
\input{fig_ap_hist_par}%
\caption{Performance (AP) per query on \rpar~+~\r1m with Medium setup. AP is shown with a bar for 8 methods. The methods, from left to right, are {\color{ba}{\protect\contour{bA}{\hesaff--\rsift--\asmk+\sp}}}, {\color{bb}{\protect\contour{bB}{\delf--\asmk+\sp}}}, {\color{bc}{\protect\contour{bC}{\delf--\hqe+\sp}}}, {\color{bd}{\protect\contour{bD}{V--[O]--R-MAC}}}, {\color{be}{\protect\contour{bE}{R--[O]--GeM}}}, {\color{bf}{\protect\contour{bF}{R--\cite{RTC17}--GeM}}}, {\color{bg}{\protect\contour{bG}{R--\cite{RTC17}--GeM+\dfs}}}, {\color{bh}{\protect\contour{bH}{\hesaff--\rsift--\asmk+\sp $\rightarrow$ R--\cite{RTC17}--GeM+\dfs}}}. The total number of easy and hard images is printed on each histogram. Best viewed in color. \label{fig:ap_hist_par}}%
\vspace{-10pt}%
\end{figure*}

\paragraph{The best of both worlds.}
The new dataset and protocols reveal space for improvement by CNN-based global descriptors in cases where local features are still better.
Diffusion performs similarity propagation by starting from the query's nearest neighbors according to the CNN global descriptor.
This inevitably includes false positives, especially in the case of few relevant images.
On the other hand, local features, \eg with \asmk+\sp, offer a verified list of relevant images.
Starting the diffusion process from geometrically verified images obtained by BoW methods combines the benefits of the two worlds.
This combined approach, shown at the bottom part of Table~\ref{tab:sota_all}, improves the performance and supports the message that both worlds have their own benefits.
Of course this experiment is expensive and we perform it to merely show a possible direction to improve CNN global descriptors.
There are more methods that combine CNNs and local features~\cite{ZWW+16}, but we focus on the results related to methods included in our evaluation.

\section{Conclusions}
\label{sec:conclusions}
We have revisited two of the most established image retrieval datasets, that were perceived as performance saturated. To make it suitable for modern image retrieval benchmarking, we address drawbacks of the original annotation. This includes new annotation for both datasets that was created with an extra attention to the reliability of the ground truth, and an introduction of 1M hard distractor set.

An extensive evaluation provides a testbed for future comparisons and concludes that image retrieval is still an open problem, especially at large scale and under difficult viewing conditions.

{\small
\bibliographystyle{ieee}
\bibliography{egbib}
}

\end{document}